# An Importance Sampling Algorithm Based on Evidence Pre-propagation

Changhe Yuan and Marek J. Druzdzel
Decision Systems Laboratory
School of Information Sciences and Intelligent Systems Program
University of Pittsburgh
Pittsburgh, PA 15260
chy24@pitt.edu, marek@sis.pitt.edu

## Abstract

Precision achieved by stochastic sampling algorithms for Bayesian networks typically deteriorates in face of extremely unlikely evidence. To address this problem, we propose the *Evidence Pre-propagation Importance Sampling* algorithm (EPIS-BN), an importance sampling algorithm that computes an approximate *importance function* using two techniques: *loopy belief propagation* [19, 25] and $\epsilon$-*cutoff* heuristic [2]. We tested the performance of EPIS-BN on three large real Bayesian networks: ANDES [3], CPCS [21], and PATHFINDER [11]. We observed that on each of these networks the EPIS-BN algorithm outperforms AIS-BN [2], the current state of the art algorithm, while avoiding its costly learning stage.

## 1 Introduction

Bayesian networks model explicitly probabilistic independence relations among sets of variables. By factorizing a full joint distribution over a set of variables into a product of conditional distributions, a Bayesian network dramatically reduces the number of parameters that are required to represent this distribution. However, exact inference in Bayesian networks is still worst case NP-hard [4]. Although approximate inference to any desired precision is worst-case NP-hard as well [5], it is the only feasible alternative for sufficiently large and densely connected networks.

A prominent subclass of approximate inference algorithms are stochastic sampling algorithms. Some of these are *probabilistic logic sampling* [12], *likelihood weighting* [6, 24], *backward sampling* [7], and *importance sampling* [24]. A subclass of stochastic sampling methods, called *Markov Chain Monte Carlo (MCMC)* methods, includes *Gibbs sampling*, *Metropolis sampling*, and *Hybrid Monte Carlo sampling* [8, 10, 17]. Stochastic sampling algorithms work well in predictive inference, but for diagnostic reasoning, especially with unlikely evidence, they often fail to provide good results within limited resources. However, given a good *importance function*, importance sampling algorithms may yield excellent approximate posteriors in a reasonable time. Researchers already proposed some methods for pre-computing good *importance functions*, such as those in the AIS-BN algorithm [2], the IS algorithm [14], and the IS_T algorithm [23]. In this paper, we propose a new importance sampling algorithm, which we call *Evidence Pre-propagation Importance Sampling algorithm for Bayesian Networks* (EPIS-BN). In this algorithm, we first use *loopy belief propagation* to compute an approximation of the optimal *importance function*, and then apply $\epsilon$-*cutoff* heuristic to cut off small probabilities in the *importance function*. We test the EPIS-BN algorithm on several large real Bayesian networks and compare the results with the AIS-BN algorithm. The empirical results show that the EPIS-BN algorithm provides a considerable improvement over the AIS-BN algorithm, especially in those cases that are hard for the latter.

The outline of the paper is as follows. In Section 2, we give a general introduction to importance sampling. We also summarize the main idea of one of its variation, the AIS-BN algorithm, which is the current state of the art algorithm and with which we later compare the EPIS-BN algorithm. In Section 3, we discuss the EPIS-BN algorithm. First, we give an introduction to *loopy belief propagation* algorithm and then explain how the EPIS-BN algorithm uses it to calculate an *importance function*. After that, we present the details of the EPIS-BN algorithm. In Section 4, we describe the results of experimental tests of the EPIS-BN algorithm on several large real Bayesian networks. Finally, in Section 5, we summarize our results, and then suggest several possible further research topics.

## 2 Importance Sampling in Bayesian Networks

We feel that it is necessary to take a look at the theoretical roots of importance sampling. Let $f(X)$ be a function of $n$ variables $X = (X_1, ..., X_n)$ over domain $\Omega \subset R^n$. Consider the problem of estimating the multiple integral

$$\mathbf{I} = \int_\Omega f(X)dX . \quad (1)$$

We assume that the domain of integration of $f(X)$ is bounded, i.e., that $\mathbf{I}$ exists. Importance sampling approaches this problem by estimating

$$\mathbf{I} = \int_\Omega \frac{f(X)}{g(X)} g(X)dX , \quad (2)$$

where $g(X)$, which is called the *importance function*, is a probability density function such that $g(X) > 0$ for any $X \subset \Omega$. $g(X)$ should be easy to sample from. In order to estimate the integral, we generate samples $X_1, X_2, ..., X_N$ from $g(X)$ and use the generated values in the sample-mean formula

$$\hat{\mathbf{I}} = \frac{1}{N} \sum_{i=1}^{N} \frac{f(X_i)}{g(X_i)} . \quad (3)$$

Importance sampling assigns more weight to regions where $f(X) > g(X)$ and less weight to regions where $f(X) < g(X)$ to correctly estimate $\mathbf{I}$. It is easy to see from Eq. 2 that $\frac{f(X)}{g(X)}$ is an unbiased estimator of $\mathbf{I}$. Rubinstein [22] points out that if $f(X) > 0$, the optimal *importance function* is

$$g(X) = \frac{f(X)}{\mathbf{I}} . \quad (4)$$

However, finding $\mathbf{I}$ is equivalent to solving the integral, so the method appears useless. But if we can find a function that is close enough to the optimal *importance function*, we can still expect good convergence rate. To get a better convergence rate, it is also important, as noted by Geweke [9], that the tails of $g(X)$ do not decay faster than the tails of $\frac{f(X)}{\mathbf{I}}$. Otherwise, the convergence rate will be slow.

Since it is impossible to get the optimal *importance function*, we should set a good *importance function* as our goal. Cheng & Druzdzel [2] proposed a method to calculate such an *importance function* in the AIS-BN algorithm. Empirical results showed that the AIS-BN algorithm achieved over two orders of magnitude improvement in convergence over *likelihood weighting* and *self-importance sampling*. The improvement came mainly from two heuristics: (1) initializing the probability distributions of parents of evidence nodes to uniform distribution, and (2) adjusting very small probabilities in the conditional probability tables. In addition to these two heuristics, the AIS-BN algorithm adopts an *importance function* learning step to approach the optimal *importance function*.

Although the two heuristics are cleverly designed, they themselves do not lead to a good *importance function*, but rather accelerate the *importance function* learning step, which is rather time consuming. In addition, the learned importance function may decay faster than the tails of the optimal *importance function*. In this paper, we propose an algorithm that directly computes an approximation of the optimal *importance function* rather than learning it.

## 3 EPIS-BN: Evidence Pre-propagation Importance Sampling Algorithm

In predictive inference, since both evidence and soft evidence are in the roots of the network, stochastic sampling can easily reach high precision. However, in diagnostic reasoning, especially when the evidence is extremely unlikely, sampling algorithms can exhibit a mismatch between the sampling distribution and the posterior distribution. In such cases most samples may be incompatible with the evidence and be useless. Some stochastic sampling algorithms such as *likelihood weighting* and *importance sampling* try to make use of all the samples by assigning weights for them. But most of the weights turn out to be too small to be effective. Backward sampling [7] tries to deal with this problem by sampling backward from the evidence nodes, but it may fail to consider the soft evidence in the roots [23]. Whatever sampling order is chosen, a good *importance function* has to take into account the information ahead in the network. If we do sampling in the topological order of the network, we need an *importance function* that will match the information from the evidence nodes. In the EPIS-BN algorithm, we make use of the *loopy belief propagation* to calculate such an *importance function*.

### 3.1 Loopy Belief Propagation

The goal of the belief propagation algorithm [20] is to find the posterior beliefs of each node $X$, i.e., $BEL(x) = P(X = x|\mathbf{E})$, where $\mathbf{E}$ denotes the set of evidence. In a polytree, any node $X$ d-separates $\mathbf{E}$ into two subsets $\mathbf{E}^+$, the evidence connected to the parents of $X$, and $\mathbf{E}^-$, the evidence connected to the children of $X$. Given the state of $X$, the two subsets are independent. Therefore, node $X$ can collect messages separately from them in order to compute its posterior beliefs. The message from $E^+$ is defined as

$$\pi(x) = P(x|\mathbf{E}^+) \quad (5)$$

and the message from $E^-$ is defined as

$$\lambda(x) = P(\mathbf{E}^-|x) \,. \quad (6)$$

By decomposing $\pi(x)$ and $\lambda(x)$ into more detailed messages between neighboring nodes, we can calculate $\pi(x)$ and $\lambda(x)$ for all the nodes by propagating messages throughout the network. [20] gives the details about how to calculate the messages. After we get the messages, we can compute the posterior beliefs of X by

$$BEL(x) = \alpha \lambda(x) \pi(x) \,. \quad (7)$$

With slight modifications, we can apply Pearl's belief propagation algorithm to networks with loops. The resulting algorithm is called *loopy belief propagation* [19, 25]. In general, *loopy belief propagation* will not give the correct posteriors for networks with loops. However, recently researchers performed extensive investigations on the performance of *loopy belief propagation*, and reported surprisingly accurate results [1, 18, 19, 25]. As of now, more thorough understanding of why the results are so good has yet to be developed. For our purpose of getting an approximate *importance function*, whether or not *loopy belief propagation* converges to the correct posteriors is not critical.

### 3.2 The EPIS-BN Algorithm

Let $\mathbf{X} = \{X_1, X_2, ..., X_n\}$ be the set of variables in a Bayesian network, $PA(X_i)$ be the parents of $X_i$, $\mathbf{E}$ be the set of evidence. Based on the theoretical considerations in Section 2, we know that the optimal importance function is

$$\rho(\mathbf{X} \backslash \mathbf{E}) = P(X|\mathbf{E}) \,. \quad (8)$$

After factorizing $P(X|\mathbf{E})$, we get

$$\rho(\mathbf{X} \backslash \mathbf{E}) = \prod_{i=1}^{n} P(X_i|PA(X_i), \mathbf{E}) \,, \quad (9)$$

where each $P(X_i|PA(X_i), \mathbf{E})$ is defined as *importance conditional probability table* (ICPT) [2].

**Definition 1** *An* importance conditional probability table (ICPT) *of a node $X_i$ is a table of posterior probabilities $P(X_i|PA(X_i), \mathbf{E})$ conditional on the evidence and indexed by its immediate predecessors, $PA(X_i)$.*

The AIS-BN [2] algorithm adopts a long learning step to learn approximations of these ICPTs, and hence the *importance function*. The following theorem shows that in polytrees we can calculate them directly.

**Theorem 1** *Let $X_i$ be a variable in a polytree, and $\mathbf{E}$ be the set of evidence. The exact ICPT $P(X_i|PA(X_i), E)$ for $X_i$ is*

$$\alpha(PA(X_i))P(X_i|PA(X_i))\lambda(X_i) \,, \quad (10)$$

*where $\alpha(PA(X_i))$ is a normalizing constant dependent on $PA(X_i)$.*

*Proof:* Let $\mathbf{E} = \mathbf{E}^+ \cup \mathbf{E}^-$, where $\mathbf{E}^+$ is the evidence connected to the parents of $X_i$, and $\mathbf{E}^-$ is the evidence connected to the children of $X_i$, then

$$\begin{aligned}
&P(X_i|PA(X_i), \mathbf{E}) \\
=\ &P(X_i|PA(X_i), \mathbf{E}^+, \mathbf{E}^-) \\
=\ &P(X_i|PA(X_i), \mathbf{E}^-) \\
=\ &\frac{P(\mathbf{E}^-|X_i, PA(X_i))P(X_i|PA(X_i))}{P(\mathbf{E}^-|PA(X_i))} \\
=\ &\alpha(PA(X_i))\lambda(X_i)P(X_i|PA(X_i)) \,. \square
\end{aligned}$$

Given Theorem 1 and Eq. 9, we have the following corollary.

**Corollary 1** *For a polytree, the optimal importance function is*

$$\rho(\mathbf{X} \backslash \mathbf{E}) = \prod_{i=1}^{n} \alpha(PA(X_i))P(X_i|PA(X_i))\lambda(X_i) \,. \quad (11)$$

If a node has no descendant with evidence, its ICPT is identical to its CPT. This property is also explained in Theorem 2 in [2].

In networks with loops, getting the exact $\lambda$ messages for all variables is equivalent to calculating the exact solutions, which is an NP-hard problem. However, because our goal is to obtain a good, not necessarily optimal *importance function*, we can satisfy it by calculating approximations of the $\lambda$ messages. Given the surprisingly good performance of *loopy belief propagation*, we believe it can also provide us with good approximate $\lambda$ messages.

Another heuristic method that we use in EPIS-BN is $\epsilon$-*cutoff* [2], i.e., setting some threshold $\epsilon$ and replacing any smaller probability in the network by $\epsilon$. At the same time, we compensate for this change by subtracting it from the largest probability in the same conditional probability distribution. This method is originally used in AIS-BN to speed up its *importance function* learning step [2]. We use it for a different purpose. Since we use approximate messages to calculate the *importance function*, we are likely to violate the requirement that the tails of our *importance function* do not decay faster than the optimal *importance function*. We try to satisfy this requirement by adjusting the small probabilities in our ICPTs. However, the optimal threshold value is highly network dependent. Furthermore, if the calculated importance function already satisfies this requirement, we may get worse *importance function* if we still apply $\epsilon$-*cutoff*.

---

1. Order the nodes according to their topological order.

2. Initialize parmaters $m$ (number of samples), $\epsilon$ and $d$ (propagation length).

3. Initialize the messages that all evidence nodes send to themselves to be vectors of a 1 for the observed state and 0's for other states, and all other messages to be uniformly vectors of 1's.

4. **for** $i \leftarrow 1$ **to** $d$ **do**

5. 　For all of the nodes, recompute their new outgoing messages based on the incoming messages from the last iteration for all of the nodes.
   **end for**

6. Calculate the *importance function* based on the final messages.

7. Enhance the *importance function* by the $\epsilon$-*cutoff* heuristic.

8. **for** $i \leftarrow 1$ **to** $m$ **do**

9. 　$\mathbf{s}_i \leftarrow$ generate a sample according to $P(\mathbf{X}|\mathbf{E})$

10. 　Compute the importance score $w_{iScore}$ of $\mathbf{s}_i$.

11. 　Add $w_{iScore}$ to the corresponding entry of each score table.
    **end for**

12. Normalize each score table, output the estimated beliefs for each node.

---

Figure 1: The Evidence Pre-propagation Importance Sampling Algorithm for Bayesian Networks (EPIS-BN).

The basic EPIS-BN algorithm is outlined in Figure 1. The parameter $m$, the number of samples, is a matter of tradeoff between precision and time. More samples will lead to a better precision. However, the optimal values of the propagation length $d$ and the threshold value $\epsilon$ for $\epsilon$-*cutoff* are highly network dependent. We will recommend some values bases on our empirical results in Section 4.2.

## 4 Experimental Results

To test the performance of the EPIS-BN algorithm, we applied it to several large real Bayesian networks, and compared our results to those of AIS-BN, the current state of the art algorithm. This section presents the results of our experiments. We implemented our algorithm in C++ and performed our tests on a Pentium III, 733 MHz Windows XP computer.

### 4.1 Experimental Method

To compare the accuracy of sampling algorithms, we compare their departure from the exact solutions, which we calculate using the clustering algorithm [16]. The distance metric we use is Hellinger's distance [15]. Hellinger's distance between two distributions $f1$ and $f2$, which have probabilities $P_1(x_{ij})$ and $P_2(x_{ij})$ for state $j$ ($j = 1, 2, ..., n_i$) of node $i$ respectively, such that $X_i \notin \mathbf{E}$ is defined as:

$$H(F_1, F_2) = \sqrt{\frac{\sum_{X_i \in \mathbf{N}\backslash\mathbf{E}} \sum_{j=1}^{n_i} \{\sqrt{P_1(x_{ij})} - \sqrt{P_2(x_{ij})}\}^2}{\sum_{X_i \in \mathbf{N}\backslash\mathbf{E}} n_i}}, \quad (12)$$

where $\mathbf{N}$ is the set of all nodes in the network, $\mathbf{E}$ is the set of evidence nodes, and $n_i$ is the number of states for node $i$.

Hellinger's distance weights small absolute probability differences near 0 much more heavily than similar probability differences near 1. In many cases, Hellinger's distance provides results that are equivalent to Kullback-Leibler measure. However, a major advantage of Hellinger's distance is that it can handle zero probabilities, which are common in Bayesian networks. For comparison purpose, in some results we also report the *Mean Square Error* (MSE).

### 4.2 Parameter Selection

The most important tunable parameter in our algorithm is the propagation length $d$. Since we are using *loopy belief propagation* only to get the approximate $\lambda$ messages, we need not wait until *loopy belief propagation* converges. We can simply adopt a propagation length equal to the depth of the deepest evidence node. However, two problems arise here. First, usually the influence of evidence on a node attenuates as the distance of the node from the evidence becomes longer [13]. Therefore, we can save a lot of effort if we stop the propagation process after a small number of iterations. Second, for networks with loops, we would be able to avoid double counting of evidence by stopping propagation after a number of iterations that is less than the size of the smallest loop [25].

nectivity and it was shown to be difficult for the AIS-BN algorithm [2].

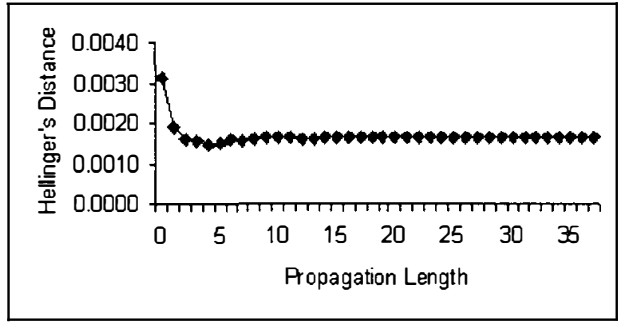

Figure 2: A plot of the influence of propagation length on the precision of the result of EPIS-BN on the ANDES Network (37 is the depth of the deepest evidence).

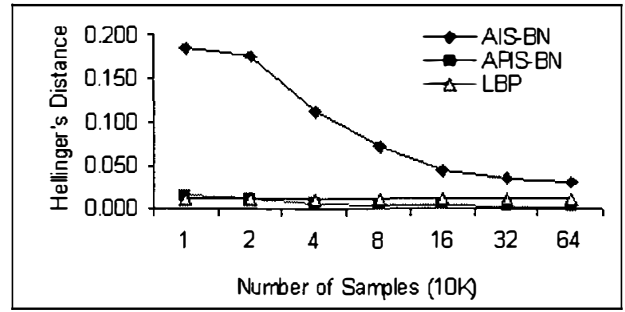

Figure 3: Convergence Rate Comparison for AIS-BN and EPIS-BN as a function of the number of samples on the ANDES network.

Figure 2 shows an experiment that we conducted to test the influence of propagation length on precision of the results in the ANDES network [3]. Other networks yielded similar results. In this case, we randomly selected 20 evidence nodes for the ANDES network. After performing different number of iterations of *loopy belief propagation*, we ran the EPIS-BN algorithm and generated $320K$ samples. The results show that a length of 2 is already sufficient to yield very good results. Increasing the propagation length to 5 improves the results minimally. Further propagation can even make the results worse. Although for different networks and evidence, the optimal propagation length was different, our experiments showed that the lengths of 4 or 5 were sufficient for deep networks. For shallow networks, we chose the depth of the deepest evidence as the propagation length.

Another important parameter in EPIS-BN is the threshold value $\epsilon$ for $\epsilon$-*cutoff*. The optimal value for $\epsilon$ is also network dependent. Our empirical tests did not yield a universally optimal value, but we recommend to use $\epsilon = 0.006$ for nodes with the number of outcomes fewer than 5, and $\epsilon = 0.001$ for nodes with the number of outcomes between 5 and 8. Otherwise, we recommend $\epsilon$ equal to 0.0005. These recommendations are different from those in [2]. The main reason for this difference is that the $\epsilon$-*cutoff* is used at a different stage of the algorithm and for a different purpose.

### 4.3 Results for the ANDES Network

The main network we used to test our algorithm on was one of the ANDES networks [3], consisting of 233 nodes. This network has a large depth and high con-

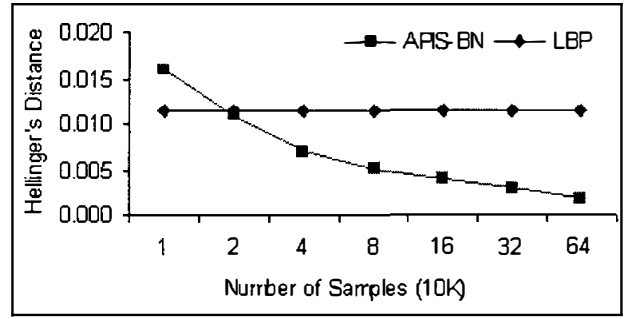

Figure 4: Convergence curve for EPIS-BN in a finer scale in Figure 3. The horizontal line shows the accuracy reached by *loopy belief propagation*.

Figure 3 shows a typical result of our experiments on the convergence rate of the AIS-BN algorithm and the EPIS-BN algorithm. Figure 4 shows the result of EPIS-BN on a finer scale. In this case, we chose the propagation length to be 5 for EPIS-BN, and randomly selected 20 evidence for the ANDES network. The prior probability of evidence was, similarly to the tests performed in [2], typically between $10^{-10}$ and $10^{-40}$. We also report the results of 100 iterations of *loopy belief propagation*. The results show that EPIS-BN achieved a precision near one order of magnitude higher than AIS-BN, while AIS-BN performed even worse than *loopy belief propagation*. Our comparison was based on the number of samples. However, AIS-BN has a long learning step, which took about 7.7 seconds for the ANDES network. EPIS-BN took only 0.5 second to do belief propagation. Relative to the sampling time, which was about 30 seconds when the number of samples is 320K, they really made a difference. If we take into account this time discrepancy, EPIS-BN can achieve even better performance within the same time. In the tables below, we do not include these times.

|  | EPIS(H) | AIS(H) | EPIS(M) | AIS(M) |
|---|---|---|---|---|
| $\mu$ | 0.0029 | 0.0590 | 0.0034 | 0.0739 |
| $\sigma$ | 0.0012 | 0.0504 | 0.0014 | 0.0636 |
| min | 0.0010 | 0.0021 | 0.0012 | 0.0025 |
| median | 0.0026 | 0.0437 | 0.0031 | 0.0560 |
| max | 0.0065 | 0.2188 | 0.0079 | 0.2776 |

Table 1: Summary of the simulation results for all the 75 simulation cases on the ANDES network. H stands for Hellinger's distance, and M for MSE.

We generated a total of 75 test cases on the ANDES network. These cases consisted of five sequences of 15 cases each. For each sequence, we randomly chose a different number of evidence nodes: 15, 20, 25, 30, 35 respectively. The evidence nodes were chosen from a predefined list of potential evidence nodes. In each test case, we set the propagation length to be 5, and ran both EPIS-BN and AIS-BN on the network for $320K$ samples. Table 1 summarizes these 75 test cases. It shows that EPIS-BN was significantly better than AIS-BN. The results of a paired one-tailed t-test for Hellinger's distance and MSE are $3.24E-15$ and $4.01E-15$ respectively. They show us highly significant difference between EPIS-BN and AIS-BN on the ANDES network.

### 4.4 Results for Other Networks

In addition to the ANDES network, we also tested EPIS-BN on two other networks, CPCS [21] and PATHFINDER [11]. Although the results were not as spectacular as those on the ANDES network, we still observed improvement.

The CPCS (Computer-based patient Case Study) network is a model representing a subset of the domain of internal medicine. It has many small probabilities, typically on the order of $10^{-4}$. The version we used had 179 variables, a subset of the full version.

|  | EPIS(H) | AIS(H) | EPIS(M) | AIS(M) |
|---|---|---|---|---|
| $\mu$ | 0.00080 | 0.00097 | 0.00060 | 0.00076 |
| $\sigma$ | 0.00020 | 0.00046 | 0.00016 | 0.00034 |
| min | 0.00062 | 0.00064 | 0.00030 | 0.00037 |
| median | 0.00073 | 0.00084 | 0.00057 | 0.00068 |
| max | 0.00183 | 0.00386 | 0.00142 | 0.00255 |

Table 2: Summary of the simulation results for all the 75 simulation cases on the CPCS network. H stands for Hellinger's distance, and M for MSE.

We also generated 75 test cases on the CPCS network by the same experiment. The only difference is that since the CPCS network is a relatively shallow network, we dynamically set the propagation length to be the depth of the deepest evidence. Most of the potential evidence nodes are leaf nodes in the network. Here also, the prior probability of evidence was extremely small, between $10^{-10}$ and $10^{-40}$ with a median of $10^{-25}$. The learning overhead of AIS-BN was 5.4 seconds, while the *loopy belief propagation* took only 0.4 second for EPIS-BN. The sampling step cost about 25 seconds. Table 2 summarizes the 75 test cases on the CPCS network.

|  | EPIS(H) | AIS(H) | EPIS(M) | AIS(M) |
|---|---|---|---|---|
| $\mu$ | 0.00072 | 0.00077 | 0.00036 | 0.00039 |
| $\sigma$ | 0.00021 | 0.00033 | 0.00011 | 0.00014 |
| min | 0.00034 | 0.00039 | 0.00021 | 0.00020 |
| median | 0.00067 | 0.00070 | 0.00034 | 0.00037 |
| max | 0.00178 | 0.00263 | 0.00076 | 0.00121 |

Table 3: Summary of the simulation results for all the 75 simulation cases on the PATHFINDER network. H stands for Hellinger's distance, and M for MSE.

The third large real network that we used in our tests is the PATHFINDER network [11]. The version we used consists of 135 nodes. Since the PATHFINDER network has many probabilities that are equal to 1 and 0, we hit zero probability evidence sometimes when generating a test case. So we ran the experiment multiple times, and collected the first 75 effective test cases, in which the generated evidence had non-zero probability. The learning overhead for AIS-BN was 3.5 second, while only 0.3 second for EPIS-BN. The sampling step cost about 15 seconds. Table 3 summarizes the 75 test cases for the PATHFINDER network.

The improvement of the EPIS-BN algorithm over the AIS-BN algorithm for the CPCS network and the PATHFINDER network is smaller than that for the ANDES network. To test whether this smaller difference is due to the ceiling effect, we performed experiments on these networks without evidence. When no evidence is present, both EPIS-BN and AIS-BN reduce to *probabilistic logic sampling* [12]. We ran *probabilistic logic sampling* on all three networks with the same number of samples as in the main experiment. We observed that the precision of the results was in the order of $10^{-4}$ (for both measures). Because when no evidence is present, the importance function is the ideal importance function, it is reasonable to say that $10^{-4}$ is the best precision that a sampling algorithms can achieve given the same resources. In case of the CPCS and the PATHFINDER networks, AIS-BN already comes very close to this precision. Therefore, the improvement of EPIS-BN over AIS-BN in the CPCS network and the PATHFINDER network is sig-nificant.

### 4.5 The Role of Loopy Belief Propagation and $\epsilon$-cutoff in EPIS-BN

Since EPIS-BN is based on *loopy belief propagation* (P) in combination with the $\epsilon$-cutoff heuristic (C), we performed experiments that aimed at disambiguating their role. We denote EPIS-BN without any heuristic method as the E algorithm. E + PC represents the EPIS-BN algorithm. We compared the performance of E, E+P, E+C, E+PC. We tested these algorithms on the same test cases generated in the previous experiments. The results are given in Table 4. The results show that the performance improvement is coming mainly from *loopy belief propagation*. The $\epsilon$-cutoff heuristic demonstrated inconsistent performance. For the CPCS and PATHFINDER networks, it helped to achieve a better precision, while it made the precision worse for the ANDES network. We believe that there are at least two explanations of this observation. First, the ANDES network has a much deeper structure than the other two networks. The loops in the ANDES network are also much larger. *Loopy belief propagation* performs much better in networks with this kind of structure. After belief propagation, the network already has near optimal ICPTs. There is no need to apply $\epsilon$-cutoff heuristic any more. Second, the proportion of small probabilities in these networks is different. The ANDES network only has 5.8 percent small probabilities, while the CPCS network has 14.1 percent and the PATHFINDER has 9.5 percent. More extreme probabilities will make the inference task more difficult, so $\epsilon$-cutoff plays a more important role in the CPCS and PATHFINDER networks. Nevertheless, the role of the $\epsilon$-cutoff heuristic still needs to be understood better.

## 5 Conclusion

It is widely believed that unlikely non-root evidence nodes and extremely small probabilities in Bayesian networks are the two main stumbling blocks for stochastic sampling algorithms. The EPIS-BN algorithm tries to overcome them by applying *loopy belief propagation* to calculate an *importance function*. Thus, we are able to take into account the influence of non-root evidence beforehand when we do sampling in the topological order in a network. The second technique, the $\epsilon$-cutoff heuristic, was originally proposed in [2], and it amounts to cutting off smaller probabilities by some threshold. This heuristic helps the tails of the *importance function* not to decay faster than the optimal *importance function*. The resulting algorithm is elegant in the sense of focusing clearly on pre-computing the importance function without a costly learning stage. Our experimental results show that the EPIS-BN algorithm achieves a considerable improvement over the AIS-BN algorithm, especially in cases that were difficult for the latter. Experimental results also show that the improvement comes mainly from *loopy belief propagation*. As the performance of the EPIS-BN algorithm will depend on the degree to which *loopy belief propagation* will approximate the posterior probabilities, techniques to avoid oscillations in *loopy belief propagation* may lead to some performance improvements.

|   |   | E | E+P | E+C | E+PC |
|---|---|---|---|---|---|
| A | $\mu$ | 0.0234 | 0.0027 | 0.0505 | 0.0029 |
| N | $\sigma$ | 0.0222 | 0.0011 | 0.0412 | 0.0012 |
| D | min | 0.0013 | 0.0010 | 0.0033 | 0.0010 |
| E | median | 0.0183 | 0.0026 | 0.0410 | 0.0026 |
| S | max | 0.1456 | 0.0060 | 0.1892 | 0.0065 |
| C | $\mu$ | 0.16077 | 0.00131 | 0.08224 | 0.00080 |
| P | $\sigma$ | 0.09987 | 0.00271 | 0.06228 | 0.00020 |
| C | min | 0.00144 | 0.00067 | 0.00181 | 0.00062 |
| S | median | 0.15575 | 0.00083 | 0.06947 | 0.00073 |
|   | max | 0.38577 | 0.02258 | 0.29699 | 0.00183 |
| P | $\mu$ | 0.07482 | 0.00088 | 0.02557 | 0.00072 |
| A | $\sigma$ | 0.09781 | 0.00026 | 0.03216 | 0.00021 |
| T | min | 0.00090 | 0.00031 | 0.00187 | 0.00034 |
| H | median | 0.03317 | 0.00083 | 0.01299 | 0.00067 |
|   | max | 0.50156 | 0.00223 | 0.22629 | 0.00178 |

Table 4: Summary of the simulation results for different algorithms in the ANDES, CPCS, PATHFINDER networks.